\documentclass{article}

\PassOptionsToPackage{dvipsnames,table}{xcolor}

\usepackage{microtype}
\usepackage{graphicx}
\usepackage{subcaption}
\usepackage{booktabs}

\usepackage{hyperref}

\usepackage[accepted]{icml2026}

\usepackage{amsmath}
\usepackage{url}
\usepackage{multirow}
\usepackage[normalem]{ulem}
\usepackage{makecell}
\usepackage{xcolor}
\usepackage{threeparttable}
\usepackage[most]{tcolorbox}
\usepackage{courier}
\usepackage{colortbl}

\definecolor{darkblue}{rgb}{0, 0, 0.5}
\hypersetup{colorlinks=true, citecolor=darkblue, linkcolor=darkblue, urlcolor=darkblue}

\newif\ifshowcomments
\showcommentstrue  

\ifshowcomments
  \newcommand{\vincent}[1]{\textcolor{violet}{[Vincent: #1]}}
  \newcommand{\todo}[1]{\textcolor{red}{[TODO: #1]}}
  \newcommand{\dawn}[1]{\textcolor{purple}{[Dawn: #1]}}
  \newcommand{\chenguang}[1]{\textcolor{orange}{[Chenguang: #1]}}
  \newcommand{\xy}[1]{\textcolor{brown}{[Xiaoyuan: #1]}}
\else
  \newcommand{\yuqi}[1]{}
  \newcommand{\vincent}[1]{}
  \newcommand{\todo}[1]{}
  \newcommand{\dawn}[1]{}
  \newcommand{\chenguang}[1]{}
  \newcommand{\xy}[1]{}
\fi

\icmltitlerunning{Controlling Tool Use with Heading-Specific Activation Steering}

\begin{document}

\twocolumn[
  \icmltitle{Controlling Tool Use with Heading-Specific Activation Steering}


  \icmlsetsymbol{equal}{*}

  \begin{icmlauthorlist}
    \icmlauthor{Yuqi Chen}{ucsc}
    \icmlauthor{Vincent Siu}{ucsc}
    \icmlauthor{Yang Liu}{ucsc}
    \icmlauthor{Dawn Song}{ucb}
    \icmlauthor{Chenguang Wang}{ucsc}
  \end{icmlauthorlist}

  \icmlaffiliation{ucsc}{Department of Computer Science and Engineering, UC Santa Cruz}
  \icmlaffiliation{ucb}{Department of Computer Science, UC Berkeley}

  \icmlcorrespondingauthor{Chenguang Wang}{chenguangwang@ucsc.edu}

  \icmlkeywords{Activation Steering, Tool Use, Large Language Models, Representation Engineering, ICML}

  \vskip 0.3in
]

\printAffiliationsAndNotice{}  

\begin{abstract}
\label{abstract}
Tool-augmented large language models extend their capabilities beyond parametric knowledge through external tools, but tend to invoke them unnecessarily. We investigate whether tool-use decisions have any stable internal representation that can be extracted and manipulated, a question that is non-trivial given that tools exist entirely in context at inference time and have no direct encoding in model weights. We show that steering vectors extracted from heading-anchors positions exert bidirectional causal control over tool-invocation behavior across five open-source models and three domains, suppressing unnecessary tool use most effectively in domains where parametric reasoning suffices. However, geometric analysis reveals that this causal effectiveness does not correspond to clean linear structure: tool-invocation steps exhibit diffuse, bimodal alignment with the suppression vector rather than the consistent negative alignment a linear encoding account would predict, and different tool types recruit largely distinct internal signatures with low cross-tool feature overlap. We hypothesize these geometric properties are indicative of the non-parametric nature of tools, and distinguish tool-use steering vectors from those extracted for parametrically grounded concepts. The relationship between this geometric irregularity and the observed causal effectiveness remains an open question.
\end{abstract}


\section{Introduction}
\label{introduction}

Tool-augmented LLMs have become a dominant paradigm for complex reasoning and real-world tasks \citep{yao2023reactsynergizingreasoningacting, st_webagentbench, schick2023toolformer}. External tools such as web search, code execution, and user-interaction modules extend model capabilities beyond parametric knowledge, improving performance on tasks requiring up-to-date information, precise computation, or clarification of underspecified intent \citep{schick2023toolformer, agentai_survey}.

However, tool augmentation also introduces a persistent failure mode: \emph{tool overuse}. Models may invoke tools even when internal reasoning would suffice, repeat calls without meaningful progress, or rely on tools in ways that increase latency and cost while providing limited benefit \citep{qian-etal-2025-smart, shen-etal-2024-smartcal}. Unnecessary tool calls can expose the system to noisy retrieval, execution failures, and avoidable error propagation \citep{wang-etal-2025-self}. Existing mitigations either require expensive retraining \citep{qian-etal-2025-smart, shen-etal-2024-smartcal} or operate at the output interface rather than on the internal activations underlying tool selection \citep{wang-etal-2025-self}, leaving the latent decision to invoke a tool unaddressed.

We ask whether tool-use decisions have any stable internal representation that can be extracted and manipulated. This is non-obvious: unlike concepts such as sentiment or factuality that are encoded in model weights through training, tools exist entirely in context at inference time, and standard assumptions from the linear representation hypothesis~\citep{park2024linearrepresentationhypothesisgeometry} and representation engineering~\citep{zou2023representationengineeringtopdownapproach} do not straightforwardly apply to concepts with no parametric grounding. Prior steering work has targeted concepts with clear parametric grounding (sentiment, refusal, truthfulness) \citep{panickssery2024steeringllama2contrastive, siu2025steeringsafetysystematicsafetyevaluation}; we are among the first to apply it to a behavior that is explicitly non-parametric by construction. Following contrastive activation addition~\citep{turner2024steeringlanguagemodelsactivation, panickssery2024steeringllama2contrastive}, we extract hidden states at structured tool-heading positions in reasoning trajectories to build tool-specific steering vectors, and evaluate their causal and geometric properties across five open-source instruction-tuned models and three domains.

\begin{figure*}[t]
    \centering
    \includegraphics[width=\textwidth]{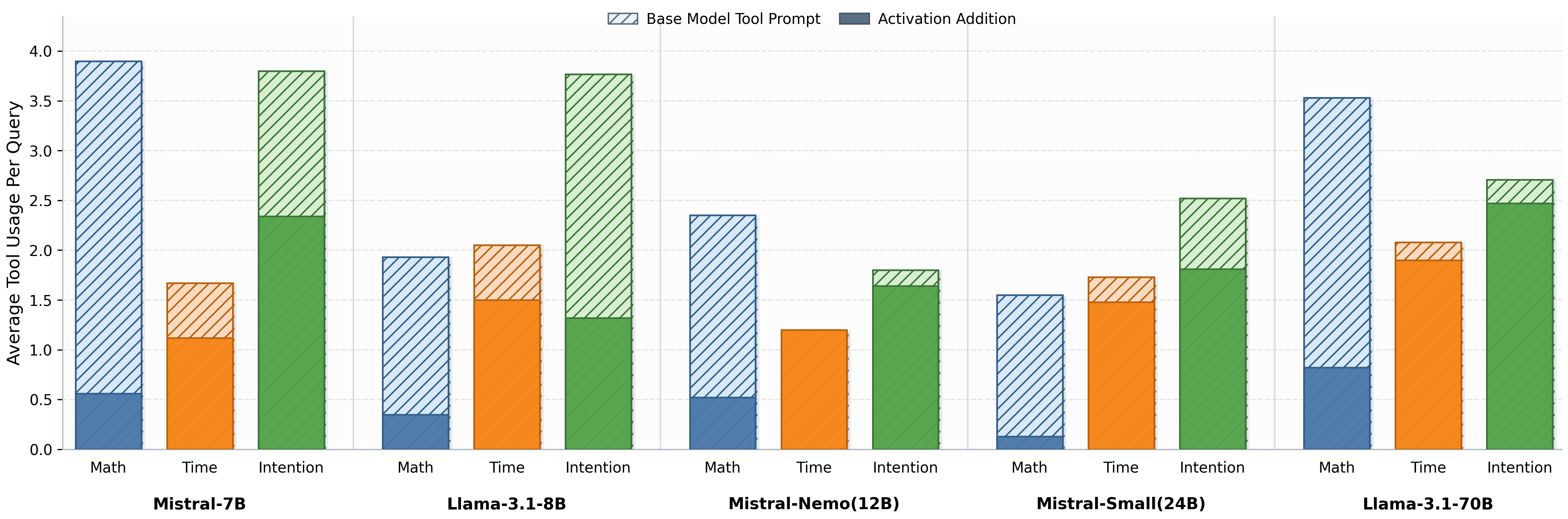}
    \caption{Average tool calls per query across five models under three conditions. Base Model Tool Prompt is the prompt-only baseline without hidden-state intervention. Activation Addition subtracts the steering vector from the residual stream, suppressing tool use below the baseline. Orthogonalization projects hidden states to be orthogonal to the steering vector, removing the suppressive component and amplifying tool use above the baseline. The opposing directions of the two interventions confirm that the extracted vector encodes a causally effective, bidirectional tool-use direction in the residual stream.}
    \label{fig:tool_usage}
\end{figure*}

Applying these vectors via activation addition suppresses tool use substantially below baseline, while orthogonalizing hidden states with respect to them amplifies tool use above it. This bidirectional behavioral control, shown in Figure~\ref{fig:tool_usage}, confirms that the extracted vectors are causally relevant to tool-use decisions despite the absence of any stable parametric encoding of tools in model weights. Suppression is most effective in domains where parametric reasoning suffices, and degrades where external retrieval or user clarification is genuinely required, with larger models showing greater resistance across all domains.

However, geometric analysis reveals that causal effectiveness does not correspond to clean linear structure. Tool-invocation steps exhibit a diffuse bimodal cosine similarity distribution with respect to the suppression vector rather than the consistent negative alignment a linear encoding account would predict, and different tool types recruit largely distinct internal feature dimensions with low cross-tool overlap. These properties are consistent with the non-parametric nature of tools and distinguish tool-use steering vectors from those extracted for parametrically grounded concepts. The relationship between this geometric irregularity and the observed causal effectiveness remains an open question, and we discuss it in the context of the multi-turn dependency structure of tool-augmented trajectories that limits direct reconciliation with the probing analysis.

\section{Related Work}
\label{Related Work}

\paragraph{Representation Theory.} The linear representation hypothesis~\citep{park2024linearrepresentationhypothesisgeometry} predicts that concepts encoded in model weights should be linearly decodable from activations, with supporting evidence from word vector arithmetic~\citep{mikolov2013linguistic} and superposition theory~\citep{elhage2022toymodelssuperposition}. Empirical work on refusal~\citep{arditi2024refusallanguagemodelsmediated} and truthfulness~\citep{burns2024discoveringlatentknowledgelanguage} has found steering vectors with clean geometric properties consistent with this account. More recent work has complicated the picture: concept directions can span multi-dimensional subspaces rather than single directions~\citep{wollschlager2025geometryrefusallargelanguage, siu2025repitrepresentingisolatedtargets}, and their geometry can exhibit complex dependencies on context and layer depth. Our work extends this line of inquiry to a setting where linear representability is not expected to hold by construction: because tools are injected through context rather than encoded in weights, the geometric irregularity we observe is a predicted consequence of the non-parametric nature of tools rather than an anomaly, and it distinguishes tool-use steering vectors from those studied in prior representation theory work.
    
\paragraph{Activation Steering.} Steering methods identify directions in activation space corresponding to target behaviors and modulate them via vector addition or orthogonalization~\citep{zou2023representationengineeringtopdownapproach, arditi2024refusallanguagemodelsmediated, turner2024steeringlanguagemodelsactivation, panickssery2024steeringllama2contrastive, siu2025steeringsafetysystematicsafetyevaluation}. Directions are commonly extracted from contrastive data pairs~\citep{burns2024discoveringlatentknowledgelanguage, arditi2024refusallanguagemodelsmediated, zou2023representationengineeringtopdownapproach} and have been used for both behavior elicitation and concept removal~\citep{ravfogel2020nulloutguardingprotected, cosmic}. Inference-time steering uses probes or classifiers~\citep{li2023inference, lee2025programmingrefusalconditionalactivation} to conditionally apply interventions during forward passes. Prior steering work has largely assumed that the concept being steered has stable parametric grounding; our work applies these methods to a concept that explicitly lacks it, and examines the consequences for both causal effectiveness and geometric structure.

\textbf{LLM Tool Use.} Tool-use methods extend language models with external functions, APIs, and environments, enabling access to up-to-date information, specialized computation, and domain-specific expertise beyond parametric memory \citep{schick2023toolformer, yao2023reactsynergizingreasoningacting, qin2023toolllmfacilitatinglargelanguage, Qu_2025, shen2024llmtoolssurvey}. Prior work studies how models decide \emph{when} to call tools, \emph{which} tools to select, and \emph{how} to incorporate tool outputs into subsequent reasoning \citep{qian2024toolinklinkingtoolkitcreation}. Recent work further improves tool-use robustness through tool creation, external module integration, and alignment for more efficient tool calling under uncertainty or knowledge-boundary awareness \citep{qian2024creatortoolcreationdisentangling, liu2025toolacewinningpointsllm, qian2024investigateconsolidateexploitgeneralstrategyintertask, xu-etal-2025-alignment}. At the same time, dedicated evaluations show that models still struggle with unnecessary or incorrect tool calls, motivating benchmarks for deciding whether tools are needed and which tools should be used \citep{huang2024metatoolbenchmarklargelanguage, ning2024wtuevalwhetherornottoolusage}. Closely related to these efforts, recent work studies tool overuse and adaptive tool calling based on model uncertainty or self-awareness \citep{wang-etal-2025-self, shen-etal-2024-smartcal, qian-etal-2025-smart}. In contrast to methods that explicitly optimize external tool interaction policies, our work focuses on inducing the desired behavior through inference-time manipulation of internal representations.

\section{Methodology}
\label{method}
We study inference-time suppression of tool calls in structured tool-augmented generation. In our setting, the model emits intermediate reasoning using explicit section Heading-Anchors such as \texttt{\#\#\# Reasoning}, \texttt{\#\#\# Search}, \texttt{\#\#\# Code}, \texttt{\#\#\# AskUser}, and \texttt{\#\#\# Final Response}. Rather than steering the model globally throughout generation, we intervene locally at heading-formation steps, where the decoder is about to select the next section label. This yields a lightweight test-time intervention that requires no parameter updates.

During one inference run, we fix a target steering payload and a predefined intervention layer $l^\star$. The method then has three parts: (i) construct heading-anchored steering vectors from complete trajectories, (ii) detect imminent Heading-Anchors formation \texttt{\#\#\#}, and (iii) rerun only the triggered decoding step with a local hook-based suppression at layer $l^\star$.

\subsection{Heading-Anchored Steering Vectors}
\label{sec:vector_construction}

For each tool type $k \in \{\texttt{Code}, \texttt{Search}, \texttt{AskUser}\}$, we collect a set of complete labeled multi-turn trajectories with Heading-Anchors:
\[
\mathcal{D}_k = \{(x_i, y_i)\}_{i=1}^{N_k},
\]
where $x_i$ is the prompt and $y_i$ is the full structured response containing Heading-Anchors, tool calls and the corresponding tool outputs. We run the model on the full prompt-response sequence and extract the hidden states from layer $l^\star$. As a result, each Heading-Anchor representation is computed in the full left-to-right context induced by the preceding reasoning and tool interactions.



We represent each heading occurrence by the hidden state at the token corresponding to its leading \texttt{\#\#\#} anchor marker. Let $R_i$ denote the set of anchor positions for all occurrences of \texttt{\#\#\# Reasoning} in trajectory $y_i$, and let $T_i^k$ denote the set of anchor positions for all occurrences of the target heading $k$. For each layer $l$, we first average Heading-Anchor states within each trajectory:
\begin{equation}
\bar{h}_{R,i}^{(l)} = \frac{1}{|R_i|} \sum_{t \in R_i} h_{i,t}^{(l)},
\qquad
\bar{h}_{k,i}^{(l)} = \frac{1}{|T_i^k|} \sum_{t \in T_i^k} h_{i,t}^{(l)}
\end{equation}
We then form a sample-level steering delta
\begin{equation}
\Delta_{k,i}^{(l)} = \bar{h}_{R,i}^{(l)} - \bar{h}_{k,i}^{(l)},
\end{equation}
 And average these deltas across multiple trajectories:
\begin{equation}
v_k^{(l)} = \frac{1}{N_k} \sum_{i=1}^{N_k} \Delta_{k,i}^{(l)}.
\end{equation}
Under this definition, $v_k^{(l)}$ points from the \texttt{\#\#\# Reasoning} state to the corresponding target tool-heading state. Adding $v_k^{(l^\star)}$ therefore acts as a suppressive intervention, shifting the hidden state away from the tool-heading representation.

This construction is intentionally performed at the sample level rather than by pooling all heading occurrences across the dataset before taking a single global difference of means. We do so because trajectories can contain different numbers of \texttt{\#\#\# Reasoning} and tool Heading-Anchors. A pooled estimator would therefore overweight trajectories with many anchor occurrences, whereas our formulation gives each trajectory a more comparable contribution to the final steering vector.

\subsection{Activation Addition and Orthogonalization}

We then utilize the obtained steering vector $v_{k}^{(l)}$ in two different ways. At a specific layer $l*$ on the Heading-Anchor tokens \texttt{\#\#\#}, we perform activation addition \citep{turner2024steeringlanguagemodelsactivation, panickssery2024steeringllama2contrastive,siu2025steeringsafetysystematicsafetyevaluation}, altering the layer input $h_t^{(l^\star)}$ as following to suppress tool use with steering coefficient $\alpha$. In our experiments we set $\alpha = 1$.

\begin{equation}
\tilde{h}_t^{(l^\star)} = h_t^{(l^\star)} + \alpha v_k^{(l^\star)}.
\end{equation}

We also test orthogonalizing the hidden states to the vector \citep{arditi2024refusallanguagemodelsmediated, cosmic} on this Heading-Anchor tokens, to study steering to encourage tool use by removing the suppression vector.

\begin{equation}
\tilde{h}_t^{(l^\star)} = h_t^{(l^\star)} - \operatorname{proj}_{v_k^{(l^\star)}}\!\left(h_t^{(l^\star)}\right), \quad \operatorname{proj}_{v}(h) = \frac{\langle h, v \rangle}{\|v\|_2^2} v
\end{equation}

\section{Experimental Setup}
\label{experiment}
\useunder{\uline}{\ul}{}

\textbf{Tasks and Domains.} We evaluate on the Math, Time, and Intention tasks from the SMART benchmark~\citep{qian-etal-2025-smart}. These three domains expose complementary tool-use regimes: Math contains many cases where parametric reasoning suffices without external computation, Time requires access to up-to-date information unavailable in model weights, and Intention requires recovering missing user-specific details through clarification. Together they test whether a suppressive intervention can reduce unnecessary tool calls without blocking those that are genuinely required.

\textbf{Models and Baselines.} We evaluate five open-source instruction-tuned models: Mistral-7B, Llama-3.1-8B, Mistral-Nemo (12B), Mistral-Small (24B), and Llama-3.1-70B~\citep{jiang2023mistral7b, grattafiori2024llama3herdmodels}. We compare against the base model with tool prompt from ~\citet{qian-etal-2025-smart}, which provides the model with all available tools and their descriptions and allows it to decide independently whether and when to invoke them, following an iterative reasoning-then-execution loop until a final answer is produced.

\begin{table*}[t]
\centering
\setlength{\tabcolsep}{8pt}
\renewcommand{\arraystretch}{2.0}
\resizebox{\textwidth}{!}{
\begin{tabular}{llccccccc}
\hline
\multirow{2}{*}{\textbf{Method}} & \multirow{2}{*}{\textbf{Model}}
& \multicolumn{2}{c}{\textbf{Math}}
& \multicolumn{2}{c}{\textbf{Time}}
& \multicolumn{3}{c}{\textbf{Intention}} \\ \cline{3-9}
& & \textbf{ACC} & \textbf{ToolAvgUse}
& \textbf{ACC} & \textbf{ToolAvgUse}
& \makecell{\textbf{Missing Details} \\ \textbf{Recovery}}
& \makecell{\textbf{Summarized} \\ \textbf{Intention}}
& \textbf{ToolAvgUse} \\ \hline

\multirow{5}{*}{\makecell{Base Model\\Tool Prompt}}
& Mistral-7B         & \textbf{13.3} & 3.90 & \textbf{49.0} & 1.67 & \textbf{48.84 / 21.70} & \textbf{63.0} & 3.80 \\
& Llama-3.1-8B       & \textbf{51.0} & 1.93 & \textbf{56.0} & 2.05 & \textbf{54.76 / 25.90} & \textbf{70.2} & 3.77 \\
& Mistral-Nemo(12B)  & \textbf{46.0} & 2.35 & \textbf{59.0} & \textbf{1.19} & \textbf{31.35 / 5.82} & 59.3 & 1.80 \\
& Mistral-Small(24B) & \textbf{76.0} & 1.55 & \textbf{62.0} & 1.73 & \textbf{45.74 / 33.62} & \textbf{78.2} & 2.52 \\
& Llama-3.1-70B      & \textbf{67.5} & 3.53 & \textbf{63.0} & 2.08 & \textbf{45.74 / 35.96} & \textbf{61.7} & 2.71 \\ \hline

\multirow{5}{*}{Orthogonalization}
& Mistral-7B         & 11.8 & 4.82 & 40.0 & 2.92 & 13.92 / 3.18 & 58.0 & 4.82 \\
& Llama-3.1-8B       & 48.8 & 2.25 & 44.0 & 3.87 & 15.48 / 3.55 & 59.3 & 4.20 \\
& Mistral-Nemo(12B)  & 39.8 & 2.66 & 45.0 & 1.87 & 24.86 / 5.12 & 56.1 & 3.05 \\
& Mistral-Small(24B) & 71.0 & 2.21 & 51.0 & 3.20 & 28.94 / 5.63 & 66.2 & 3.62 \\
& Llama-3.1-70B      & 60.5 & 4.31 & 48.0 & 2.99 & 31.42 / 6.21 & 56.0 & 3.59 \\ \hline

\multirow{5}{*}{\makecell{Activation\\Addition}}
& Mistral-7B         & 11.8 & \textbf{0.56} & 39.0 & \textbf{1.12} & 14.3 / 4.67 & 57.6 & \textbf{2.34} \\
& Llama-3.1-8B       & 48.3 & \textbf{0.35} & 45.0 & \textbf{1.50} & 16.28 / 6.38 & 55.2 & \textbf{1.32} \\
& Mistral-Nemo(12B)  & 43.5 & \textbf{0.52} & 47.0 & 1.20 & 27.36 / 5.58 & \textbf{60.8} & \textbf{1.64} \\
& Mistral-Small(24B) & 65.4 & \textbf{0.13} & 49.0 & \textbf{1.48} & 31.05 / 5.94 & 69.4 & \textbf{1.81} \\
& Llama-3.1-70B      & 63.2 & \textbf{0.82} & 49.0 & \textbf{1.90} & 33.21 / 7.68 & 55.9 & \textbf{2.47} \\ \hline
\end{tabular}
}

\caption{
Performance comparison across three methods and three domains: \textbf{Math}, \textbf{Time}, and \textbf{Intention}.
For \textbf{Math} and \textbf{Time}, we report task accuracy (ACC) and average tool usage (ToolAvgUse).
For \textbf{Intention}, we report Missing Details Recovery, Summarized Intention, and average tool usage.
Results are grouped by model to enable fair comparison across methods.
The best value is marked in \textbf{bold}.
}
\label{mainexp}
\end{table*}

\textbf{Prompts.} Since the method described in \textbf{Method} relies on heading-anchors, we use different prompts compared to ~\citet{qian-etal-2025-smart}. For each domain, we use the prompt templates as shown in \ref{prompt template}. Specifically,we instruct the model to format each turn with an explicit heading anchor, such that every response must begin with \texttt{\#\#\#} followed by one of the predefined action types: \texttt{Reasoning, Code, Search, AskUser, Final Response}.  In this form, it standardizes the intermediate structure of model outputs across domains, making the interaction trace easier to parse and analyze.

\textbf{Steering Vector Construction.}
To construct steering vectors, we sample data from the training split of SMART benchmark~\citep{qian-etal-2025-smart}. Specifically, we select 20 examples for each of the three domains used in our experiments: \textbf{Math}, \textbf{Time}, and \textbf{Intention}, yielding 60 examples in total. These examples are used only to generate structured trajectories for steering-vector extraction. Using the prompts above, we generate structured responses for the selected SMART training examples with \textbf{GPT-4o} \citep{openai_gpt4_2023}. These generated trajectories serve as the data for computing steering vectors described in Section~\ref{sec:vector_construction}. In particular, the explicit heading structure enables us to identify the hidden states corresponding to \texttt{\#\#\# Reasoning} and the target tool heading-anchors, from which we compute domain-relevant steering directions.

\textbf{Steering Configuration.}
As described in Section~\ref{sec:vector_construction}, we construct a separate steering vector for each tool type. In the main experiments, we restrict steering to the primary tool associated with each domain: \textbf{Code} for \textbf{Math}, \textbf{Search} for \textbf{Time}, and \textbf{AskUser} for \textbf{Intention}.

\textbf{Evaluation Metrics.} Following~\cite{qian-etal-2025-smart}, for Math and Time we report task accuracy (\textbf{ACC}) and average tool calls per example (\textbf{ToolAvgUse}). For Intention we additionally report \textbf{Missing Details Recovery} and \textbf{Summarized Intention} from the IN3 dataset~\cite{qian2024tellmoreimplicituser}, alongside the overall task score and \textbf{ToolAvgUse}.

\begin{figure*}[t]
    \centering
    \includegraphics[width=\textwidth]{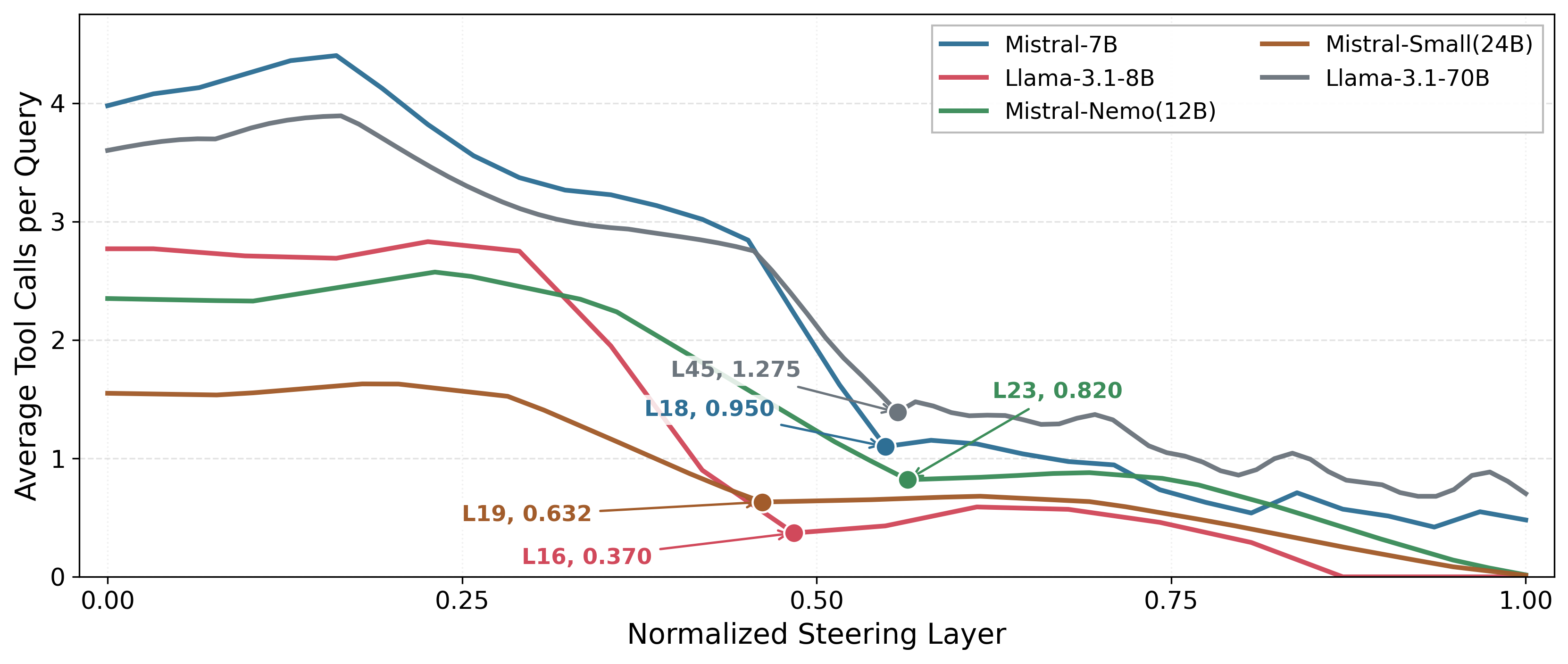}
    \caption{Average tool calls per query as a function of normalized steering layer depth across five models. Tool usage drops sharply as the steering layer moves into the middle of the network before plateauing, forming a clear elbow. Annotated points mark the selected elbow layer for each model; despite differing in absolute depth, all five fall within normalized depths 0.45--0.60, suggesting that tool-use representations stabilize at a surprisingly consistent relative depth across architectures.}
    \label{fig:layer}
\end{figure*}

\section{Steering Results}
\label{sec:steeringresults}

\textbf{Steering Layer Selection} We sweep the steering vector across all layers, performing activation addition at each layer with coefficient $\alpha = 1$ to minimize tool usage, and select the elbow of the resulting tool-usage curve (Figure~\ref{fig:layer}). Mistral-7B uses layer 18 with 0.95 average tool calls per prompt, Llama-3.1-8B uses layer 16 with 0.70, Mistral-Nemo uses layer 23 with 0.82, Mistral-Small uses layer 19 with 0.63, and Llama-3.1-70B uses layer 45 with 1.28. Despite differing in absolute depth, all five elbow points fall between normalized depths $0.45$--$0.60$, suggesting that tool-use representations stabilize at a consistent relative depth across tested models. The steering layer is calibrated once per model and held fixed for all subsequent experiments.

\textbf{Tool Use Steering Results}
Table~\ref{mainexp} presents the main experimental results across the three domains, while Figure~\ref{fig:tool_usage} visualizes the corresponding changes in average tool usage. The central pattern is consistent across all five models: activation addition strongly suppresses tool invocation, whereas orthogonalization increases it above the base-model level. This bidirectional effect is itself non-trivial: tool use is not a clean parametrically encoded skill in model weights, with limited theoretical underpinnings on linear representability. The fact that addition and orthogonalization move behavior in opposite directions suggests that the extracted vector captures a relevant component of the model's tool-use decision.

Table~\ref{mainexp} also shows that reduced tool use does not translate into uniform task outcomes across domains. In \textbf{Math}, activation addition drives tool usage close to zero for all models (e.g., from 3.90 to 0.56 for Mistral-7B, from 1.93 to 0.35 for Llama-3.1-8B, and from 1.55 to 0.13 for Mistral-Small), while accuracy drops only modestly relative to the base model in most cases. This suggests that many tool calls in Math are indeed redundant, and that parametric reasoning can often replace external computation. By contrast, in \textbf{Time} and \textbf{Intention}, suppressing tool use is consistently accompanied by larger performance degradation. For example, on Time, Llama-3.1-8B drops from 56.0 to 45.0 ACC as tool usage falls from 2.05 to 1.50, and Mistral-Small drops from 62.0 to 49.0 as tool usage falls from 1.73 to 1.48. On Intention, activation addition substantially reduces Missing Details Recovery and Summarized Intention scores across all models, indicating that many of these tool calls are not redundant but necessary for recovering underspecified user information. This finding is relatively intuitive: the questions in Time and Intention are by construction unanswerable from parametric knowledge alone, so suppressing tool calls intuitively leads to larger performance drops.


\begin{figure*}[t]
    \centering
    \includegraphics[width=\textwidth]{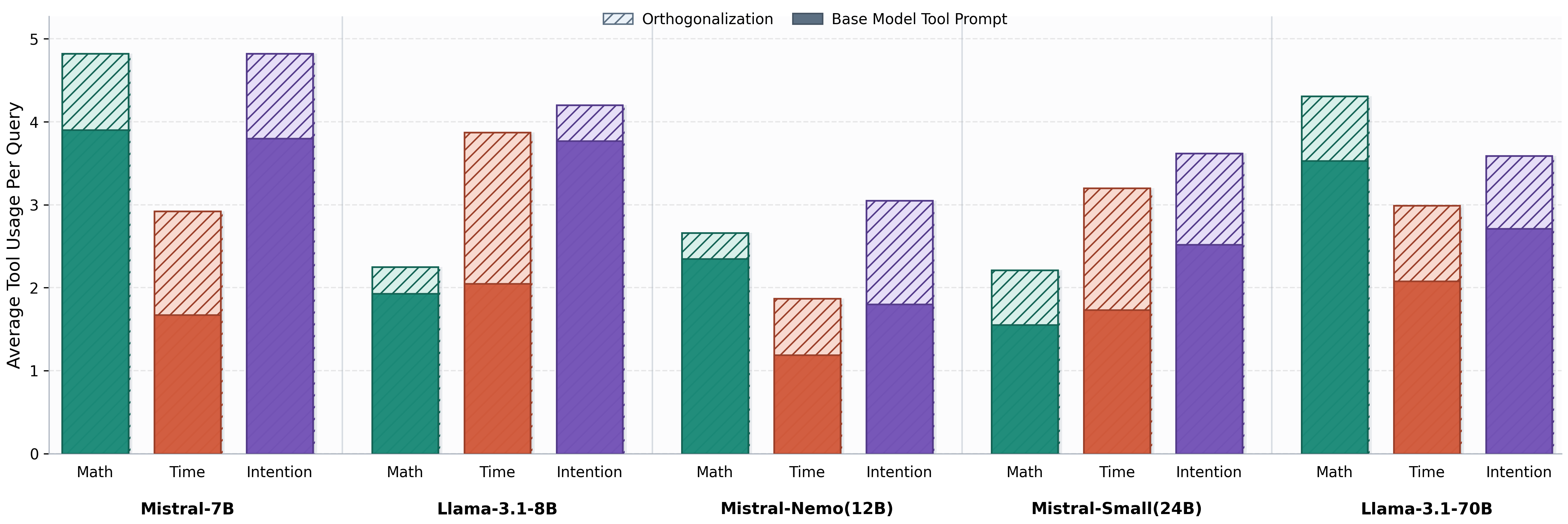}
    \caption{Average tool calls per query under orthogonalization and the base model prompt across five models and three domains. Orthogonalization projects hidden states to be orthogonal to the suppression vector at tool-heading positions, removing its suppressive component. Tool usage increases above the base model level in all cases, suggesting that the suppression vector is active in the unsteered baseline and that its removal releases tool-invocation behavior.}
    \label{fig:orthogonalization}
\end{figure*}

Figure~\ref{fig:tool_usage} makes this pattern visually clear: activation addition suppresses tool use across all models and domains, while orthogonalization shifts behavior in the opposite direction. This domain asymmetry is strongest in Math, where tool calls are nearly eliminated, suggesting that many invocations are redundant. By contrast, Time and Intention remain more resistant to suppression, consistent with these domains relying more heavily on retrieval or clarification that cannot be replaced by parametric knowledge alone.

Orthogonalization provides the complementary intervention. As shown in Table~\ref{mainexp} and Figure~\ref{fig:orthogonalization}, projecting hidden states to be orthogonal to the suppression direction increases tool usage above the base model across all models and domains. However, this additional tool use does not improve overall task performance; in most settings it instead lowers accuracy or intention quality relative to the base model. This indicates that the base model is not simply under-using tools. Rather, it already reflects a partial internal balance between tool invocation and direct answering, and removing the suppressive component pushes the model toward overuse.
This suggests that steering tool use may entangle with parametric knowledge due to the non-parametric nature of tools, a phenomenon also observed in prior representation-steering work \citep{siu2025steeringsafetysystematicsafetyevaluation}. Taken together, the results suggest that the extracted direction primarily controls \emph{whether} the model invokes tools, but the utility of suppressing that behavior depends strongly on domain: it is comparatively safe in Math, but much more harmful in Time and Intention, where tool access is often genuinely required.

\textbf{Robustness Beyond the Literal Heading Token.}
Because the intervention is anchored on the formatted heading \texttt{\#\#\# Code}, a natural concern is whether it controls tool invocation or merely suppresses the literal surface string. We probe this with two controls on Llama-3.1-8B in the Math domain (full results in Appendix~\ref{app:heading_robustness}). First, under \emph{cross-format transfer}, we extract the \texttt{Code} vector under the Markdown heading schema but evaluate it under a JSON action schema (e.g.\ \texttt{\{"action": "code", \ldots\}}) in which the literal \texttt{\#\#\# Code} heading never appears; the Markdown-extracted vector still removes $63.9\%$ of tool calls with accuracy essentially unchanged ($0.330 \to 0.325$). Second, under \emph{heading rename}, we keep the vector extracted from \texttt{\#\#\# Code} but evaluate with renamed tool headings of the same semantics; suppression persists even when the active heading contains no occurrence of the word \texttt{Code} (e.g.\ $70.3\%$ relative suppression under \texttt{\#\#\# Action\_B}). Both controls indicate that the effect is not solely the suppression of the literal heading token. It is nonetheless schema-sensitive: reverse transfer (JSON$\to$Markdown) is much weaker ($15.5\%$), and JSON$\to$JSON reduces tool use but collapses accuracy through malformed generations. We therefore characterize the extracted direction as a schema- and anchor-sensitive tool-use control direction with partial transfer beyond literal heading tokens, rather than a fully format-independent representation. We note these are preliminary single-model, single-domain controls.

\begin{figure*}[!t]
    \centering
    \includegraphics[width=\textwidth]{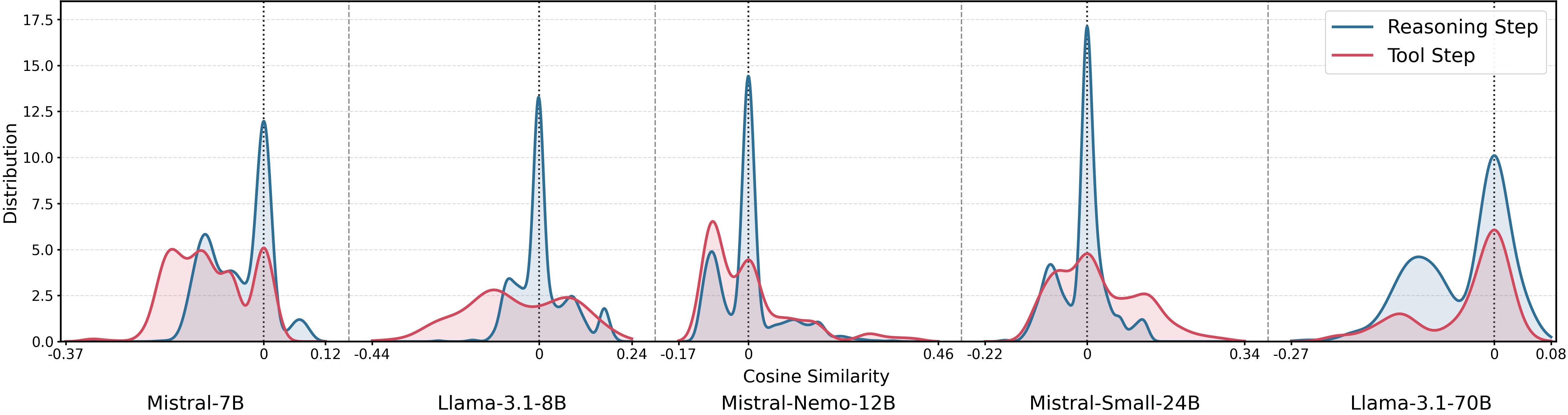}
    \caption{Cosine similarity between the steering vector and hidden states at reasoning steps (blue) and tool-invocation steps (red) on unsteered prompts, across five models. Reasoning steps concentrate sharply at zero, indicating near-orthogonality to the steering vector. Tool steps exhibit a diffuse bimodal distribution with mass on both the acute and oblique sides, inconsistent with the steering vector acting as a simple linear encoding of tool-use intent. This pattern is consistent across all five tested models. However, we note that due to the multi-turn nature of the evaluation, these similarities do not describe the responses intervened on in Section~\ref{sec:steeringresults}, since steering shifts the trajectory of each interaction.}
    \label{fig:cossim_distributions}
\end{figure*}

\section{Geometric Analysis of the Steering Vector}
\label{sec:cosine}

\begin{figure*}[t]
    \centering
    \includegraphics[width=\textwidth]{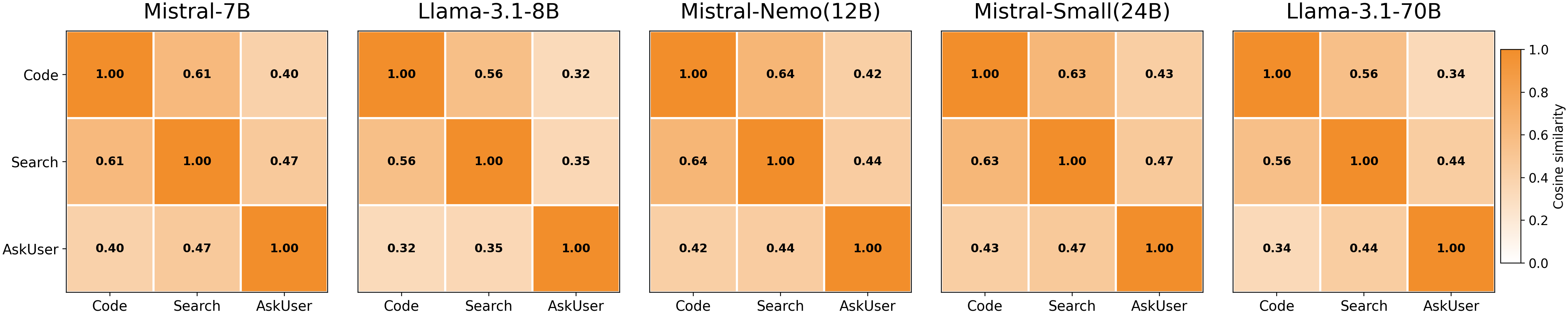}
    \caption{Pairwise cosine similarity between steering vectors extracted for code, search, and askuser tool types across five models. Code and search vectors are consistently more similar to each other than either is to the askuser vector, suggesting that clarification-seeking recruits distinct internal structure relative to external computation and retrieval.}
    \label{fig:cosine_heatmaps}
\end{figure*}

\begin{table*}[t]
\centering
\caption{Jaccard similarity of the top 10\% feature-dimension indices selected by absolute steering-vector magnitude for the \texttt{Code}, \texttt{Search}, and \texttt{AskUser} steering vectors across models. Each value reports the mean overlap across vectors; the last column shows the three-way Jaccard overlap among all three vectors.}
\label{tab:steeringvectorsim}
\renewcommand{\arraystretch}{1.2}
\begin{tabular}{lcccc}
\hline
Model & Code\&Search & Code\&AskUser & Search\&AskUser & Three-way \\
\hline
Mistral-7B         & 0.198 & 0.143 & 0.179 & 0.0671 \\
Llama-3.1-8B       & 0.181 & 0.117 & 0.127 & 0.0416 \\
Mistral-Nemo(12B)  & 0.226 & 0.144 & 0.165 & 0.0639 \\
Mistral-Small(24B) & 0.228 & 0.149 & 0.170 & 0.0675 \\
Llama-3.1-70B      & 0.178 & 0.108 & 0.140 & 0.0421 \\
\hline
\end{tabular}
\end{table*}

\subsection{Probing Expression of the Steering Vectors}

To understand what the steering vector captures geometrically, we measure the cosine similarity between the extracted vector and the hidden states of unsteered prompts at both reasoning steps and tool-invocation steps. If the vector encoded a stable tool-suppression direction, tool-step hidden states should align negatively with it while reasoning-step hidden states should not. Importantly, these distributions are characterized on unsteered multi-turn trajectories, where each step is conditioned on prior tool interactions; the geometric relationship between the suppression vector and the hidden state distribution under orthogonalization in Figure~\ref{fig:orthogonalization} is therefore not directly comparable, as orthogonalization intervenes on a sequential process in which each modified step conditions all subsequent ones.

Figure~\ref{fig:cossim_distributions} plots these distributions across all five models. Reasoning steps concentrate sharply at zero, indicating near-orthogonality to the steering vector across all architectures. Tool steps exhibit a generally negative alignment, but their distributions are often diffuse and bimodal, with substantial mass in both acute and oblique alignment with the steering vector. This is seemingly consistent with the non-parametric nature of tools: because tool definitions exist only in context at inference time and have no direct encoding in model weights, tool-step activations have no a priori reason to cluster along a consistent direction in the residual stream. The extracted vector therefore does not correspond to a stable concept direction in the sense assumed by prior representation engineering work~\citep{zou2023representationengineeringtopdownapproach, panickssery2024steeringllama2contrastive}, but rather captures something about the conditions under which the model defers to context. We hypothesize that steering effectiveness persists despite this geometric irregularity because the heading-anchor format concentrates the tool-invocation decision into a single token: intervening at this structural bottleneck may be sufficient for behavioral control even when the underlying concept has no consistent direction in the residual stream.

In Llama-3.1-70B this geometric irregularity is most pronounced: the tool-step hidden states approach near-orthogonality to the mean-difference steering vector, making them difficult to distinguish from reasoning-step states under this vector alone. A key question is whether this reflects an \emph{absence} of any linearly encoded heading-type information at scale, or merely a limitation of the mean-difference estimator. To disentangle the two, we compare the mean-difference direction against alternative extractors on Llama-3.1-70B---a supervised linear probe, a diagonally whitened mean difference, and norm-matched random and shuffled-label controls---each scored for both linear separability and causal steering effect (Appendix~\ref{app:extraction_70b}, Table~\ref{tab:extraction_70b}). The results show that heading-type information \emph{is} linearly decodable at 70B: a linear probe separates \texttt{Code} from \texttt{Reasoning} heading states almost perfectly (validation AUROC $1.000$), so we do not claim that the model lacks a linearly readable encoding of heading type. Crucially, however, this most-decodable direction is \emph{not} the most causally effective one. The mean-difference vector suppresses tool use far more strongly than the probe direction (held-out Math ToolAvgUse $0.235$ vs.\ $1.235$), the whitened estimator does not improve over it, and the random and shuffled controls are substantially weaker on both axes \citep{wu2025axbenchsteeringllmssimple, panickssery2024steeringllama2contrastive}. We therefore revise the interpretation accordingly: rather than reading the 70B geometry as evidence of linear \emph{inseparability} or of a failed estimator, we conclude that probing separability and causal controllability \emph{diverge} at this scale. The heading-type concept is linearly readable, yet the direction that most effectively \emph{controls} tool invocation remains geometrically near-orthogonal to the tool-step states it acts on. This dissociation is itself consistent with the non-parametric account: causal leverage at the heading-anchor bottleneck does not require the steering direction to coincide with the cleanest decoding direction, and the cosine analysis on unsteered multi-turn trajectories does not by itself explain the orthogonalization behavior.

\subsection{Cross-Tool Steering Vector Similarity}

The steering vectors extracted for each tool type encode which feature dimensions of the residual stream are most strongly implicated in the decision to invoke that tool. To assess whether different tools draw on similar internal representations, we examine both the pairwise cosine similarity between steering vectors (Figure~\ref{fig:cosine_heatmaps}) and the Jaccard similarity of their top 10\% of feature dimensions by absolute magnitude (Table~\ref{tab:steeringvectorsim}). High overlap under either measure would indicate that different tool invocations recruit similar regions of the model's feature space; low overlap would indicate largely distinct internal signatures.

Both measures are consistent. Cosine similarities between steering vectors range from $0.324$ to $0.643$, with code-search similarity the highest pairwise value across all five models and askuser the most dissimilar from both. Jaccard overlaps follow the same pattern: pairwise overlaps range from $0.108$ to $0.228$, with three-way overlap substantially lower at $0.042$--$0.068$. The higher code-search similarity is consistent with both involving external computation over well-defined inputs, whereas askuser involves a fundamentally different pragmatic action directed at the user rather than an external system. The low three-way overlap confirms that tool-type specificity is the dominant structure, with little shared substrate across all three vectors.

\section{Conclusion}
\label{conclusion}

This paper investigates tool-use decisions in language models as an activation steering problem: tools are non-parametric, existing entirely in context at inference time with no direct encoding in model weights. Despite this, heading-anchored steering vectors exert bidirectional causal control over tool-invocation behavior across five models and three domains, suppressing tool use via activation addition and amplifying it via orthogonalization. Suppression is most effective in domains where parametric reasoning suffices, and degrades where external retrieval or clarification is genuinely required.

Geometric analysis reveals that this causal effectiveness does not correspond to clean linear structure. Tool-invocation steps exhibit a diffuse bimodal cosine similarity distribution with respect to the suppression vector, and different tool types recruit largely distinct feature dimensions with low cross-tool overlap, distinguishing these vectors from those extracted for parametrically grounded concepts. The mechanism by which causal control persists despite geometric irregularity remains unresolved; the multi-turn dependency structure of tool-augmented trajectories further limits reconciliation of the orthogonalization results with the probing analysis. More broadly, these findings suggest that the boundary between parametric and non-parametric knowledge is less clean than behavioral evidence suggests, and that context-injected behaviors may be more amenable to representation-level intervention than their non-parametric nature would predict.

\bibliography{main}
\bibliographystyle{icml2026}

\newpage
\appendix
\onecolumn
\section{Appendix}

\subsection{Prompts}
\label{prompt template}
\newtcolorbox{promptbox}[1][]{
    colback=gray!5,
    colframe=black,
    fonttitle=\bfseries,
    title=#1,
    sharp corners,
    boxrule=0.8pt,
    left=6pt,
    right=6pt,
    top=6pt,
    bottom=6pt,
}

\begin{promptbox}[Math-Domain Initial Prompt]

\begin{ttfamily}
\#\#\# Task\\
You are a highly capable assistant designed to solve tasks effectively using your knowledge and the tool set available for this domain.\\

\#\#\# Principles\\
1. Reason Independently:\\
- Leverage your own knowledge to analyze and solve reasoning steps whenever possible. Use the tool only when necessary.\\
2. Tool Usage:\\
- Use code snippet ```python ... ``` inside a `\#\#\# Code` step when computation is needed.\\
3. Step-by-Step Approach:\\
- Work through reasoning systematically, breaking down the task into manageable steps. Rely on your knowledge until a gap is identified that requires tool support.\\
- Use only the domain-appropriate tool when needed.\\
4. Goal-Oriented Resolution:\\
- Conclude your reasoning process by achieving a clear, accurate, and succinct solution based on your independent analysis and any tool findings.\\

\#\#\# Output Guidelines\\
- Your answer must begin with `\#\#\# Reasoning`.\\
- After that, every step must begin with one of these section headers on its own line: `\#\#\# Reasoning`, `\#\#\# Code`, `\#\#\# Final Response`.\\
- If you need the tool, place the executable snippet inside a `\#\#\# Code` step using ```python ... ```.\\
- Do not emit any other tool headings.\\
- After a tool call, continue reasoning using the tool output when available.\\
- End with a `\#\#\# Final Response` step that directly answers the task.
\#\#\# Task:\\
\end{ttfamily}
\end{promptbox}

\begin{promptbox}[Time/Intention-Domain Initial Prompt]
\label{time intention prompt}
\begin{ttfamily}
\#\#\# Task\\
You are a highly capable assistant designed to solve tasks effectively using your knowledge and the tool set available for this domain.\\

\#\#\# Principles\\
1. Reason Independently:\\
- Leverage your own knowledge to analyze and solve reasoning steps whenever possible. Use the tool only when necessary.\\
2. Tool Usage:\\
- Use a `\#\#\# Search` step whose content is a concise web search query when external information is needed.\\
- Use a `\#\#\# AskUser` step whose content is the exact question that should be asked to the user when user-specific clarification is needed.\\
3. Step-by-Step Approach:\\
- Work through reasoning systematically, breaking down the task into manageable steps. Rely on your knowledge until a gap is identified that requires tool support.\\
- Use only the domain-appropriate tool when needed.\\
4. Goal-Oriented Resolution:\\
- Conclude your reasoning process by achieving a clear, accurate, and succinct solution based on your independent analysis and any tool findings.\\

\#\#\# Output Guidelines\\
- Your answer must begin with `\#\#\# Reasoning`.\\
- After that, every step must begin with one of these section headers on its own line: `\#\#\# Reasoning`, `\#\#\# Search`, `\#\#\# AskUser`, `\#\#\# Final Response`.\\
- If you use `\#\#\# Search`, place only the search query inside the step.\\
- If you use `\#\#\# AskUser`, place only the user-facing question inside the step.\\
- Do not emit any other tool headings.\\
- After a tool call, continue reasoning using the tool output when available.\\
- End with a `\#\#\# Final Response` step that directly answers the task.\\

\#\#\# Task:\\
\end{ttfamily}

\end{promptbox}

\subsection{Robustness Beyond the Literal Heading Token}
\label{app:heading_robustness}

To test whether the steering vector controls tool invocation or merely suppresses the literal surface string \texttt{\#\#\# Code}, we run two control studies on Llama-3.1-8B in the Math domain. Relative Suppression denotes $(\text{Base}-\text{ActAdd})/\text{Base}$ on ToolAvgUse.

In the \emph{cross-format transfer} study (Table~\ref{tab:crossformat}), the \texttt{Code} vector is extracted under one output schema and evaluated under another; the \texttt{markdown}$\to$\texttt{json} row applies a Markdown-extracted vector under a JSON action schema in which the literal \texttt{\#\#\# Code} heading is absent. The vector still suppresses $63.9\%$ of tool calls with accuracy essentially unchanged, but transfer is asymmetric: the reverse \texttt{json}$\to$\texttt{markdown} direction is much weaker ($15.5\%$), and \texttt{json}$\to$\texttt{json} reduces tool use while collapsing accuracy because JSON-format generations are frequently malformed.

\begin{table}[h]
\centering
\caption{Cross-format transfer on Llama-3.1-8B (Math). The vector is extracted under the schema before the arrow and evaluated under the schema after it.}
\label{tab:crossformat}
\renewcommand{\arraystretch}{1.2}
\resizebox{\columnwidth}{!}{
\begin{tabular}{lccccc}
\hline
\multirow{2}{*}{Schema} & Base & ActAdd & Relative & Base & ActAdd \\
& ToolAvgUse & ToolAvgUse & Suppression & Accuracy & Accuracy \\
\hline
markdown $\to$ markdown & 3.150 & 0.225  & 92.9\% & 0.5075 & 0.495 \\
markdown $\to$ json     & 1.053 & 0.380  & 63.9\% & 0.330  & 0.325 \\
json $\to$ markdown     & 3.150 & 2.6625 & 15.5\% & 0.5075 & 0.5225 \\
json $\to$ json         & 1.053 & 0.300  & 72.9\% & 0.330  & 0.010 \\
\hline
\end{tabular}
}
\end{table}

In the \emph{heading rename} study (Table~\ref{tab:headingrename}), the vector is always extracted from \texttt{\#\#\# Code} trajectories but evaluation uses alternative tool headings of the same semantics. Suppression persists even when the active heading contains no occurrence of the word \texttt{Code} (e.g.\ $70.3\%$ under \texttt{\#\#\# Action\_B}) and accuracy is not degraded, though the effect weakens for some renames (e.g.\ \texttt{\#\#\# Execute}, $55.5\%$).

\begin{table}[h]
\centering
\caption{Heading rename on Llama-3.1-8B (Math). The vector is extracted from \texttt{\#\#\# Code}; evaluation uses the renamed heading after the arrow.}
\label{tab:headingrename}
\renewcommand{\arraystretch}{1.2}
\resizebox{\columnwidth}{!}{
\begin{tabular}{lccccc}
\hline
\multirow{2}{*}{Schema} & Base & ActAdd & Relative & Base & ActAdd \\
& ToolAvgUse & ToolAvgUse & Suppression & Accuracy & Accuracy \\
\hline
Code $\to$ Code     & 3.150 & 0.170 & 94.6\% & 0.5075 & 0.525  \\
Code $\to$ Compute  & 2.877 & 0.245 & 91.5\% & 0.5427 & 0.530  \\
Code $\to$ Execute  & 3.135 & 1.395 & 55.5\% & 0.5275 & 0.5475 \\
Code $\to$ Action\_B & 2.897 & 0.860 & 70.3\% & 0.525  & 0.550  \\
\hline
\end{tabular}
}
\end{table}

\subsection{Alternative Extraction Methods on Llama-3.1-70B}
\label{app:extraction_70b}

To separate linear \emph{decodability} of heading type from the \emph{causal} steering effect of the extracted direction on Llama-3.1-70B, we compare the mean-difference direction (used throughout the paper) against four alternatives at layer $45$ on the Math/Code heading pair: a supervised linear probe, a diagonally whitened mean difference, a norm-matched random direction, and a shuffled-label probe. Every direction is norm-matched to the original mean-difference vector and evaluated on $200$ held-out Math examples. Val AUROC and Val F1 measure how well the direction separates \texttt{Code} from \texttt{Reasoning} heading states; ToolAvgUse and Accuracy measure the behavioral effect of steering with that direction (lower ToolAvgUse indicates stronger suppression).

As shown in Table~\ref{tab:extraction_70b}, the linear probe separates the two classes almost perfectly (AUROC $1.000$), confirming that heading-type information is linearly decodable at 70B. Yet this most-decodable direction is far weaker as a causal control (ToolAvgUse $1.235$) than the mean-difference vector ($0.235$), and the whitened estimator does not improve over the mean difference. The random and shuffled controls are weak on both axes. Probing separability and causal controllability therefore diverge at this scale, and the mean-difference estimator is not simply a poor estimator for causal suppression at 70B.

\begin{table}[h]
\centering
\caption{Alternative extraction directions on Llama-3.1-70B (Math/Code, layer 45, $n=200$). Each direction is norm-matched to the original mean-difference vector. AUROC/F1 measure linear separability of \texttt{Code} vs.\ \texttt{Reasoning} heading states; ToolAvgUse/Accuracy measure the causal steering effect.}
\label{tab:extraction_70b}
\renewcommand{\arraystretch}{1.2}
\resizebox{\columnwidth}{!}{
\begin{tabular}{lcccc}
\hline
Direction & Val AUROC & Val F1 & ToolAvgUse & Accuracy \\
\hline
Mean difference               & 0.945 & 0.481 & 0.235 & 0.695 \\
Linear probe                  & 1.000 & 1.000 & 1.235 & 0.695 \\
Diag-whitened mean difference & 0.961 & 0.481 & 0.405 & 0.690 \\
Random                        & 0.614 & 0.361 & 1.680 & 0.740 \\
Shuffled probe                & 0.631 & 0.332 & 1.755 & 0.735 \\
\hline
\end{tabular}
}
\end{table}

\end{document}